\documentclass[11pt,a4paper]{article}
\usepackage[hyperref]{emnlp2018}
\usepackage{times}
\usepackage{latexsym}

\usepackage{url}
\usepackage{graphicx}  
\usepackage{amsmath}
\usepackage{algorithm}
\usepackage{algpseudocode}
\usepackage{amsfonts}
\usepackage{amssymb}
\usepackage{bm}
\usepackage{color}
\usepackage{pgfplots}

\newcommand{\vE}{{\bm{E}}}

\newcommand{\vs}{{\bm{s}}}
\newcommand{\vc}{{\bm{c}}}

\newcommand{\vh}{{\bm{h}}}

\newcommand{\vx}{{\bm{x}}}

\newcommand{\vy}{{\bm{y}}}
\newcommand{\vb}{{\bm{b}}}

\newcommand{\vu}{{\bm{u}}}
\newcommand{\vW}{{\bm{W}}}
\newcommand{\vtheta}{{\bm{\theta}}}

\newcommand{\softmax}{\textrm{softmax}}

\aclfinalcopy 

\title{Contextual Neural Model for Translating Bilingual Multi-Speaker Conversations}

\author{Sameen Maruf$^1$ \\
   Monash University \\
VIC, Australia \\\And
  Andr{\'e} F. T. Martins$^2$ \\
  Unbabel \& \\
Instituto de Telecomunicac\~oes\\
 Lisbon, Portugal\\
  {\tt {$^1$\{firstname.lastname\}@monash.edu}} \\
  {\tt $^2$andre.martins@unbabel.com} \\\And
  Gholamreza Haffari$^1$\\
Monash University \\
VIC, Australia \\}

\date{}

\begin{document}
\maketitle
\begin{abstract}


Recent works in neural machine translation have begun to explore document translation. However, translating online multi-speaker conversations is still an open problem. In this work, we propose the task of translating Bilingual Multi-Speaker Conversations, and explore neural architectures which exploit both source and target-side conversation histories for this task. To initiate an evaluation for this task, we introduce datasets extracted from Europarl v7 and OpenSubtitles2016. Our experiments on four language-pairs confirm the significance of leveraging conversation history, both in terms of BLEU and manual evaluation.
\end{abstract}

\section{Introduction}


Translating a conversation online is ubiquitous in real life, e.g. in the European Parliament, United Nations, and customer service chats. 
This scenario involves leveraging the conversation history in multiple languages. 
The goal of this paper is to propose and explore a simplified version of such a setting, referred to as Bilingual Multi-Speaker Machine Translation (Bi-MSMT), where speakers' turns in the conversation switch the source and target languages. We investigate neural architectures that exploit the bilingual conversation history for this scenario,  
%
%
which is a challenging problem as the history consists of utterances in both languages.



The ultimate aim of all machine translation systems for dialogue is to enable a multi-lingual conversation between multiple speakers. However, translation of such conversations is not well-explored in the literature.
Recently, there has been work focusing on using the discourse or document context to improve NMT, in an online setting, by using the past context \cite{Jean:17, Wang:17, Bawden:17, Voita:18}, and in an offline setting, using the past and future context \cite{Maruf:18}. 
In this paper, we design and evaluate a conversational Bi-MSMT model, where we incorporate the source and target-side conversation histories into a sentence-based attentional model \cite{Bahdanau:15}. 
Here, the source history comprises of sentences in the original language for both languages, and the target history consists of their corresponding translations.
We experiment with different ways of computing the source context representation for this task. Furthermore, we present an effective approach to leverage the target-side context, and also present an intuitive approach for incorporating both contexts simultaneously. 
To evaluate this task, we introduce datasets extracted from Europarl v7 and OpenSubtitles2016, containing speaker information.
Our experiments on English-French, English-Estonian, English-German and English-Russian language-pairs show improvements of +1.44, +1.16, +1.75 and +0.30 BLEU, respectively, for our best model over the context-free baseline.  
The results show the impact of conversation history on translation of bilingual multi-speaker conversations and can be used as benchmark for future work on this task.


\section{Related Work}
Our research builds upon prior work in the field of context-based language modelling and context-based machine translation.

\paragraph{Language Modelling}

There have been few works on leveraging context information for language modelling. \newcite{Ji:15} introduced Document Context Language Model (DCLM) which incorporates inter and intra-sentential contexts. \newcite{Hoang:16} make use of side information, e.g. metadata, and \newcite{Tran:16} use inter-document context to boost the performance of RNN language models.

For conversational language modelling, \newcite{Ji:04} propose a statistical multi-speaker language model (MSLM) that considers words from other speakers when predicting words from the current one. By taking the inter-speaker dependency into account using a normal trigram context, they report significant reduction in perplexity.

\paragraph{Statistical Machine Translation}

 The few SMT-based attempts to document MT are either restrictive or do not lead to significant improvements upon automatic evaluation. 
Few of these deal with specific discourse phenomena, such as resolving anaphoric pronouns \cite{Hardmeier:10} or lexical consistency of translations \cite{Garcia:17}. Others are based on a two-pass approach i.e., to improve the translations already obtained by a sentence-level model \cite{Hardmeier:12, Garcia:14}.

\paragraph{Neural Machine Translation} 
Using context-based neural models for improving online and offline NMT is a popular trend recently. \newcite{Jean:17} extend the vanilla attention-based NMT model  \cite{Bahdanau:15} by conditioning the decoder on the previous source sentence via a separate encoder and attention component. \newcite{Wang:17} generate a summary of three previous source sentences via a hierarchical RNN, which is then added as an auxiliary input to the decoder. \newcite{Bawden:17} explore various ways to exploit context from the previous sentence on the source and target-side by extending the models proposed by \newcite{Jean:17,Wang:17}. Apart from being difficult to scale
, they report deteriorated BLEU scores when using the target-side context. 

\newcite{Tu:2017} augment the vanilla NMT model with a continuous cache-like memory, along the same lines as the cache-based system for traditional document MT \cite{Gong:11}, which stores hidden representations of recently generated words as translation history. The proposed approach shows significant improvements over all baselines when translating subtitles and comparable performance for news and TED talks. Along similar lines, \newcite{Kuang:18} propose dynamic and topic caches to capture contextual information either from recently translated sentences or the entire document to model coherence for NMT. \newcite{Voita:18} introduce a context-aware NMT model in which they control and analyse the flow of information from the extended context to the translation model. They show that using the previous sentence as context their model is able to implicitly capture anaphora.

For the offline setting, \newcite{Maruf:18} incorporate the global source and target document contexts into the base NMT model via memory networks. They report significant improvements using BLEU and METEOR for the contextual model over the baseline. To the best of our knowledge, there has been no work on Multi-Speaker MT or its variation to date.



\section{Preliminaries}

\subsection{Problem Formulation}
We are given a dataset that comprises parallel conversations,  
and each conversation consists of \emph{turns}. Each turn is constituted by sentences spoken by a single speaker, denoted by $\mathbf{x}$ or $\mathbf{y}$, if the sentence is in English or Foreign language, respectively. 
The goal is to learn a model that is able to leverage the mixed-language conversation history in order to produce high quality translations. 
%

\subsection{Data}
Standard machine translation datasets 
are inappropriate for Bi-MSMT task since they are not composed of conversations or the speaker annotations are missing. In this section, we describe how we extract data from raw Europarl v7 \cite{Koehn:05} and OpenSubtitles2016\footnote{\url{http://www.opensubtitles.org/}} \cite{Lison:16} for this task\footnote{The data is publicly available at \url{https://github.com/sameenmaruf/Bi-MSMT.git}}.

\paragraph{Europarl}
The raw Europarl v7 corpus \cite{Koehn:05} contains 
\texttt{SPEAKER} and \texttt{LANGUAGE} tags where the latter indicates the language the speaker was actually using. 
The individual files are first split 
into conversations. 
The data is tokenised (using scripts by \newcite{Koehn:05}), and cleaned (headings and single token sentences removed). 
Conversations are divided into smaller ones if the number of speakers is greater than 5.\footnote{Using the conversations as is or setting a higher threshold further reduces the data due to inconsistencies in conversation/turn lengths in the source and target side.} 
The corpus is then randomly split into train/dev/test sets with respect to conversations in ratio 100:2:3. 
The English side of the corpus is set as reference, and if the language tag is absent, the source language is English, otherwise Foreign. The sentences in the source-side of the corpus are kept or swapped with those in the target-side based on this tag. 

We perform the aforementioned steps for English-French, English-Estonian and English-German, and obtain the bilingual multi-speaker corpora for the three language pairs. Before splitting into train/dev/test sets, we remove conversations with sentences having more than 100 tokens for English-French, English-German and more than 80 tokens for English-Estonian\footnote{Sentence-lengths of 100 tokens result in longer sentences than what we get for the other two language-pairs.} respectively, to limit the sentence-length for using subwords with BPE \cite{Sennrich:16}. The data statistics are given in Table~\ref{tab:data} and Appendix~\ref{app:data}\footnote{Although the extracted dataset is small but we believe it to be a realistic setting for a real-world conversation task, where reference translations are usually not readily available and expensive to obtain.}.

\setlength{\tabcolsep}{2pt}

\begin{table}[t!]
\centering
{\small
\begin{tabular}{l|c c c | c}
& \multicolumn{3}{c}{\textbf{Europarl}} & \multicolumn{1}{|c}{\textbf{Subtitles}}\\
\cline{2-5}
& \textbf{En-Fr} & \textbf{En-Et} & \textbf{En-De} & \textbf{En-Ru} \\
\hline
\hline
{\# Conversations} & {6997} & {4394} & 3582 & 23126 \\
{\# Sentences} & {246540} & {174218} & 109241 & 291516\\
\hline
\hline
\multicolumn{5}{c}{\textbf{Mean Statistics per Conversation}}\\
\hline
\# Sentences & 36.24 & 40.65 &  31.50 & 13.60 \\
\# Turns & 4.77 & 4.85 & 4.79 & 7.12 \\
Turn Length & 7.12 & 7.92 & 6.16 & 1.68 \\
\hline
\end{tabular}
}
\caption{General statistics for training set.}
\label{tab:data}
\end{table}

\paragraph{Subtitles}
There has been recent work to obtain speaker labels via automatic turn segmentation for the OpenSubtitles2016 corpus \cite{LisonMeena:2016,Wees:2016,Wang:2016}. We obtain the English side of OpenSubtitles2016 corpus annotated with speaker information by \newcite{LisonMeena:2016}.\footnote{The majority of sentences still have missing annotations \cite{LisonMeena:2016} due to changes between the original script and the actual movie or alignment problems between scripts and subtitles. As for \newcite{Wang:2016}, their publicly released data is even smaller than our En-De dataset extracted from Europarl.}  
To obtain the parallel corpus, we use the OpenSubtitles alignment links to align foreign subtitles to the annotated English ones. For each subtitle, we extract individual conversations with more than 5 sentences and at least two turns. 
Conversations with more than 30 turns are discarded. Finally, since subtitles are in a single language, we assign language tag such that the same language occurs in alternating turns. We thus obtain the Bi-MSMT corpus for English-Russian, which is then divided into training, development and test sets.

\subsection{Sentence-based attentional model}

Our base model consists of two sentence-based NMT architectures \cite{Bahdanau:15}, one for each translation direction. Each of them contains an encoder to \emph{read} the source sentence and an attentional decoder to \emph{generate} the target translation one token at a time. 

\paragraph{Encoder}
It maps each source word $x_m$ to a distributed representation $\vh_m$ which is the concatenation of the corresponding hidden states of 
two RNNs running in opposite directions over the source sentence. The forward and backward RNNs are taken to be GRUs (gated-recurrent unit; \newcite{Cho:14}) in this work.

\paragraph{Decoder}
The generation of each target word $y_n$ is conditioned on all the previously generated words $\vy_{<n}$ via the state $\vs_n$ of the decoder, and the source sentence via a \emph{dynamic} context vector $\vc_n$:

\vspace{-4mm}
{\small
\begin{eqnarray*}
y_n&\sim& \softmax(\vW_{y}\cdot \vu_n+\vb_y) \\
\vu_n&=& \tanh(\vs_n + \vW_{uc}\cdot \vc_n +\vW_{un}\cdot \vE_T[y_{n-1}])\\
\vs_n&=& \textrm{GRU}(\vs_{n-1},\vE_T[y_{n-1}],\vc_n)
\end{eqnarray*}
}\normalsize
where $\vE_T[y_{n-1}]$ is the embedding of previous target word $y_{n-1}$, and \{$\vW_{(\cdot)}$,$\vb_y$\} are the parameters. The fixed-length \emph{dynamic} context representation of the source sentence $\vc_n ~=~\sum_{m}\alpha_{nm}\vh_m$ is generated by an attention mechanism where $\bm{\alpha}$ specifies the proportion of relevant information from each word in the source sentence.



\section{Conversational Bi-MSMT Model}
\begin{figure}[t!]
 \centering
  \includegraphics[width=0.37\textwidth]{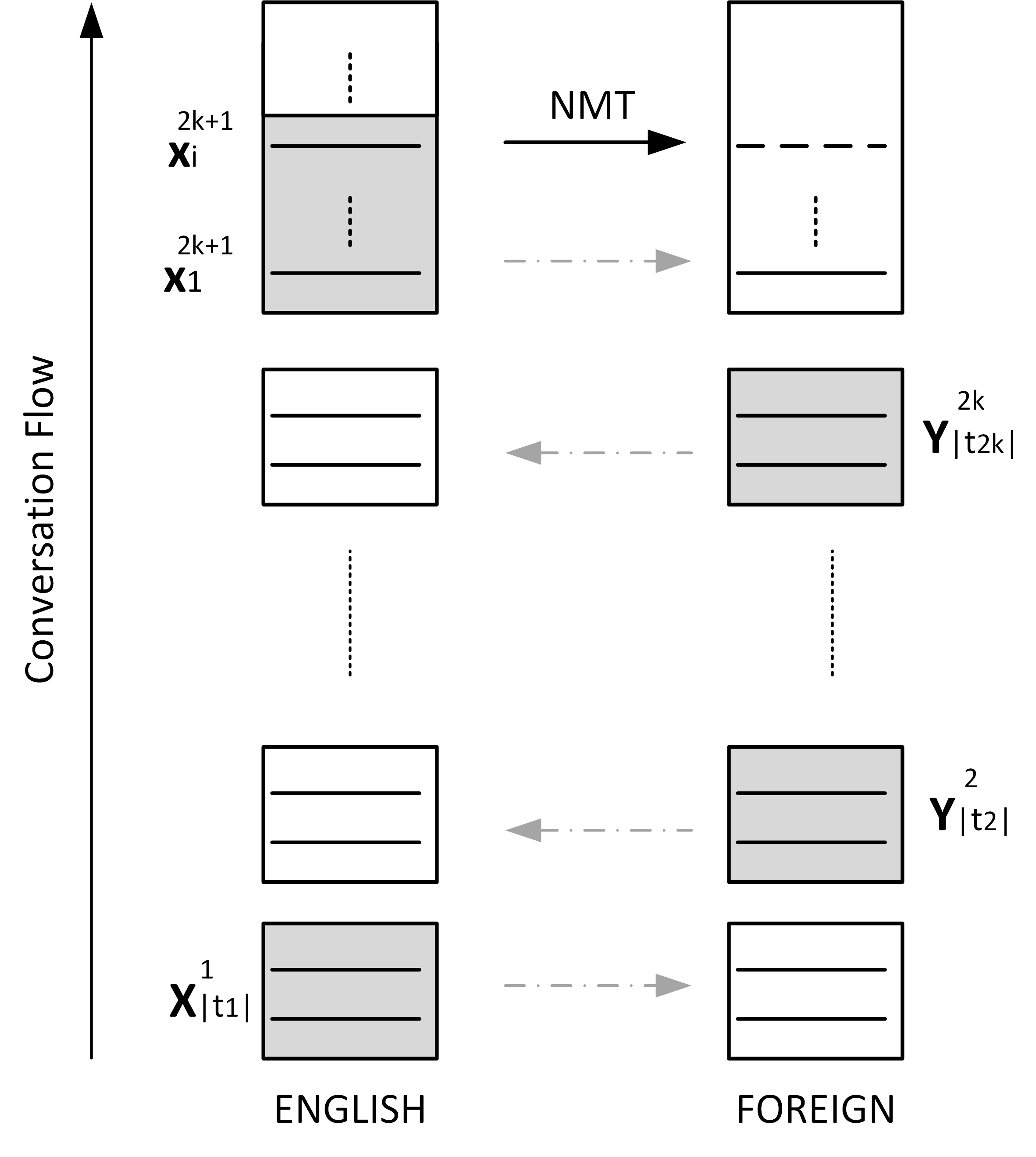}
 \caption{Overview of an ongoing conversation while translating $i^{th}$ sentence in ${2k+1}^{th}$ turn. $\mathbf{X}_{|t_j|}^j$ and $\mathbf{Y}_{|t_j|}^j$ denote the sentences in previous English and Foreign turn respectively, and $\mathbf{x}_i^j$ denotes the sentence $i$ in ongoing turn $j$ where $i\in\{1,...,|t_j|\}$. The shaded turns are observed i.e., source (the speaker utterances), while the rest are unobserved i.e., the target translations or the unuttered source sentences for current turn.}
\label{fig:overview}
\end{figure}

Before we delve into the details of how to leverage the conversation history, we identify the three types of context we may encounter in an ongoing bilingual multi-speaker conversation, as shown in Figure~\ref{fig:overview}. It comprises of: (i) the previously completed English turns, (ii) the previously completed Foreign turns, and (iii) the ongoing turn (English or Foreign). 

We propose a conversational Bi-MSMT model that is able to incorporate all three types of context using source, target or dual conversation histories into the base model. The base model caters to the speaker's language transition by having one sentence-based NMT model (described previously) for each translation direction, English$\rightarrow$Foreign and Foreign$\rightarrow$English. We now describe our approach for extracting relevant information from the source and target bilingual conversation history.

\subsection{Source-Side History}

Suppose we are translating an ongoing conversation having alternating turns of English and Foreign. We are currently in the ${2k+1}^{th}$ turn (in English) and want to translate its $i^{th}$ sentence using the source-side conversation history represented by context vector $\mathbf{o}_{src}$ (dimensions \textit{H}). 

\begin{figure}[t!]
 \centering
  \includegraphics[width=0.48\textwidth]{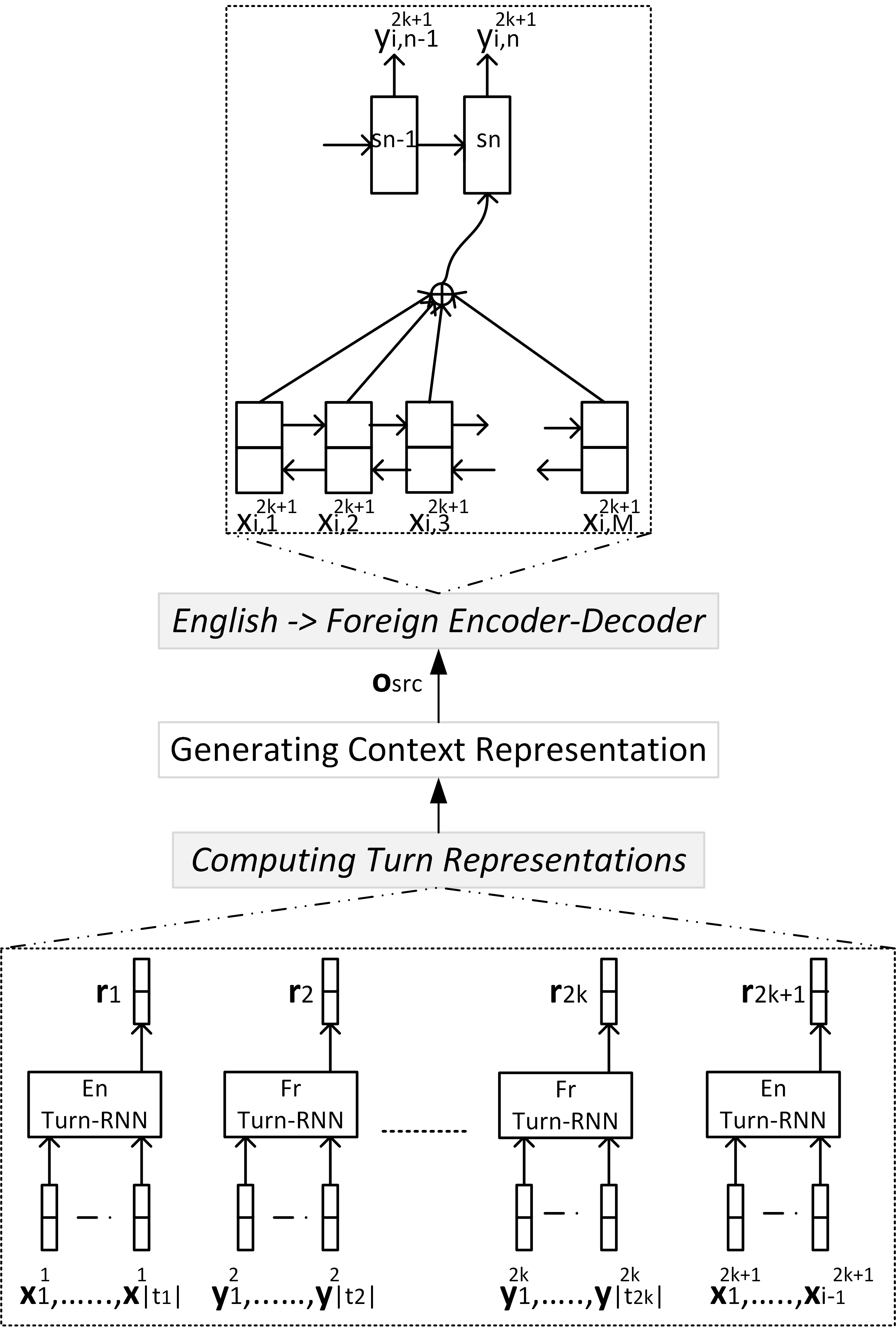}
 \caption{Architectural overview when translating $i^{th}$ sentence in ${2k+1}^{th}$ turn using source history.}
\label{fig:two-level}
\end{figure}

Let's assume that we already have the representations of previous source sentences in the conversation. We pass the source sentence representations through Turn-RNNs, which are composed of language-specific bidirectional RNNs irrespective of the speaker
, as shown in Figure~\ref{fig:two-level}, and concatenate the last hidden states of the forward and backward Turn-RNNs to get the final turn representation $\mathbf{r}_j$, where $j$ denotes the turn index. 
The individual turn representations are then combined, based on language\footnote{For this work, we define the turns based on language and do not use the speaker information as for real-world chat scenarios (e.g., agent-client in a customer service chat), we do not have multiple speakers based on language. We leave this for future exploration.}, to obtain context vectors $\mathbf{o}_{en}$ and $\mathbf{o}_{fr}$, computed in several possible ways (described below), which are further amalgamated using a gating mechanism so as to give differing importance to each element of the context vector:

\vspace{-2mm}
\begin{eqnarray}\label{eq:gate}
\mathbf{o}_{en,fr} &=& \boldsymbol{\alpha}\odot \mathbf{o}_{en}+(\mathbf{1}-\boldsymbol{\alpha})\odot \mathbf{o}_{fr}\\ 	\nonumber
\boldsymbol{\alpha} &=& \sigma(\mathbf{U}_{en}\times \mathbf{o}_{en}+\mathbf{U}_{fr}\times \mathbf{o}_{fr}+\mathbf{b}_g)
\end{eqnarray}
where $\sigma$ is the logistic sigmoid function, $\mathbf{U}$'s are matrices and $\mathbf{b}_g$ is a vector. Finally, we perform a dimensionality reduction to obtain:
\begin{equation}\label{eq:transform}
\mathbf{o}_{src}=\tanh(\mathbf{W}_T\times \mathbf{o}_{en,fr}+\mathbf{b}_T)
\end{equation}
In the remainder of this section, $\{\mathbf{W},\mathbf{U},\mathbf{b}\}$ are language-specific learned parameters. We propose five ways of computing the language-specific context representations, $\mathbf{o}_{en}$ and $\mathbf{o}_{fr}$.

\paragraph{Direct Transformation} 
The simplest approach is to combine turn representations using a language-specific dimensionality reduction transformation:

\vspace{-2mm}
{\small
\begin{eqnarray*}
\mathbf{o}_{en}&=&\tanh([\mathbf{W}_{en};...;\mathbf{W}_{en}]\times[\mathbf{r}_1;...;\mathbf{r}_{2k+1}]+\mathbf{b}_{en})\\
\mathbf{o}_{fr}&=&\tanh([\mathbf{W}_{fr};...;\mathbf{W}_{fr}]\times[\mathbf{r}_2;...;\mathbf{r}_{2k}]+\mathbf{b}_{fr})
\end{eqnarray*}
}\normalsize
Here $\mathbf{r}_j$'s are concatenated row-wise.

\paragraph{Hierarchical Gating} 
We propose a language-specific exponential decay gating based on the intuition that the farther the previous turns are from the current one, the lesser their impact may be on the translation of a sentence in an ongoing turn, similar in spirit to the caching mechanism by \newcite{Tu:2017}:

\vspace{-2mm}
{\small
\begin{eqnarray*}
\mathbf{o}_{en}=g_{en}(g_{en}(...g_{en}(g_{en}(\mathbf{r}_1,\mathbf{r}_3),\mathbf{r}_5)...),\mathbf{r}_{2k-1}),\mathbf{r}_{2k+1})
\end{eqnarray*}
}
where
\vspace{-2mm}
{\small
\begin{eqnarray*}\label{eq:hgate}
g_{en}(\mathbf{a},\mathbf{b}) &=& \boldsymbol{\alpha}\odot \mathbf{a}+(\mathbf{1}-\boldsymbol{\alpha})\odot \mathbf{b}\\		
\boldsymbol{\alpha} &=& \sigma(\mathbf{U}_{1,en}\times \mathbf{a}+\mathbf{U}_{2,en}\times \mathbf{b}+\mathbf{b}_{en})
\end{eqnarray*}
}
$\mathbf{o}_{fr}$ is computed in a similar way.

\paragraph{Language-Specific Attention} 
The English and Foreign turn representations are combined separately via attention to allow the model to focus on relevant turns in the English and the Foreign context: 

\vspace{-2mm}
{\small
\begin{eqnarray}\label{eq:lang}
\mathbf{p}_{en}&=&\softmax([\mathbf{r}_1;...;\mathbf{r}_{2k+1}]^T\times \mathbf{h}_i)\\		\nonumber
\mathbf{p}_{fr}&=&\softmax([\mathbf{r}_2;...;\mathbf{r}_{2k}]^T\times \tanh(\mathbf{W}_{en}\times\mathbf{h}_i+\mathbf{b}_{en}))\\		\nonumber
\mathbf{o}_{en}&=&\tanh(\mathbf{W}_{en}\times ([\mathbf{r}_1;...;\mathbf{r}_{2k+1}]\times \mathbf{p}_{en})+\mathbf{b}_{en})\\		\nonumber
\mathbf{o}_{fr}&=&[\mathbf{r}_2;...;\mathbf{r}_{2k}]\times \mathbf{p}_{fr}		\nonumber
\end{eqnarray}
}\normalsize
Here $\mathbf{r}_j$'s are concatenated column-wise, $\mathbf{h}_i$ is the concatenation of last hidden state of forward and backward RNNs in the encoder for current sentence $i$ in turn $2k+1$ (dimensions \textit{2H}) and \{$\mathbf{W}_{en}$, $\mathbf{b}_{en}$\} transform the language space to that of the target language. 

\paragraph{Combined Attention} 
This is a language-independent attention that merges all turn representations into one. The hypothesis here is to verify if the model actually benefits from Language-Specific attention or not.

\vspace{-2mm}
{\small
\begin{eqnarray*}
\mathbf{p}_{en,fr}&=&\softmax([\mathbf{r}_{1,en};\mathbf{r}_{2};...;\mathbf{r}_{2k+1,en}]^T\times\\ &&\qquad\qquad\tanh(\mathbf{W}_{en}\times\mathbf{h}_{i}+\mathbf{b}_{en}))\\	
\mathbf{o}_{en,fr}&=&[\mathbf{r}_{1,en};\mathbf{r}_{2};...;\mathbf{r}_{2k+1,en}]\times \mathbf{p}_{en,fr}
\end{eqnarray*}
}\normalsize
Here \small{$\mathbf{r}_{2k+1,en}=\tanh(\mathbf{W}_{en}\times\mathbf{r}_{2k+1}+\mathbf{b}_{en})$}\normalsize.

\paragraph{Language-Specific Sentence-level Attention} 
All the previous approaches for computing $\mathbf{o}_{en}$ and $\mathbf{o}_{fr}$ use a single turn-level representation. We propose to use the sentence information explicitly via a sentence-level attention to evaluate the significance of more fine-grained context in contrast to Language-Specific Attention. We first concatenate the hidden states of forward and backward Turn-RNNs for each sentence and get a matrix comprising of representations of all the previous source sentences, i.e., for English turns, we have $[\mathbf{r}_1^1;...; \mathbf{r}_{|t_1|}^1; ...; \mathbf{r}_{1}^{2k+1};...; \mathbf{r}_{i-1}^{2k+1}]$, and similarly we have another matrix for all the previous Foreign sentences
. Here, each $\mathbf{r}_i^j$ is the representation of source sentence $i$ in turn $j$ computed by the bidirectional Turn-RNN. The remaining computations are same as in Eq.~\ref{eq:lang}.

\subsection{Target-Side History}
Using target-side conversation history is as important as that of the source-side since it helps in making the translation more faithful to the target language. This becomes crucial for translating conversations where the previous turns are all in the same language.
For incorporating the target-side context, we use a sentence-level attention similar to the one described for the source-side context, i.e., for all previous English source sentences, we have a matrix $\mathbf{R}_{en}$ comprising of the corresponding target sentence representations in Foreign, 
and another matrix $\mathbf{R}_{fr}$ 
of target sentence representations (in English) for previous Foreign turns. Here each target sentence representation has dimensions \textit{H}. Then,

\vspace{-2mm}
{\small
\begin{eqnarray*}
\mathbf{p}_{en}&=&\softmax(\mathbf{R}_{en}^T \times \tanh(\mathbf{W}_{t,en}\times \mathbf{h}_i +\mathbf{b}_{t,en}))\\		\nonumber
\mathbf{p}_{fr}&=&\softmax(\mathbf{R}_{fr}^T \times (\mathbf{W}_{td,en}\times \mathbf{h}_i+\mathbf{b}_{td,en}))\\		\nonumber
\mathbf{o}_{en}&=&\mathbf{R}_{en}\times \mathbf{p}_{en}\\		\nonumber
\mathbf{o}_{fr}&=&\tanh(\mathbf{W}_{t,en}\times (\mathbf{R}_{fr}\times \mathbf{p}_{fr})+\mathbf{b}_{t,en})		\nonumber
\end{eqnarray*}
}\normalsize
where \{$\mathbf{W}_{t,en}$,$\mathbf{b}_{t,en}$\} are for dimensionality reduction and changing the language space of the query vector $\mathbf{h}_i$ and the context vector, while \{$\mathbf{W}_{td,en}$,$\mathbf{b}_{td,en}$\} are only for dimensionality reduction. 
$\mathbf{o}_{en}$ and $\mathbf{o}_{fr}$ are further combined using a gating mechanism as in Eq.~\ref{eq:gate} to obtain the final target context vector $\mathbf{o}_{tgt}$ (dimensions \textit{H}).

\subsection{Dual Conversation History}
Now that we have explained how to leverage the source and target conversation history separately, we explain how they can be utilised simultaneously. The simplest way to do this is to incorporate both context vectors $\mathbf{o}_{src}$ and $\mathbf{o}_{tgt}$ into the base model (explained in Sec~\ref{sec:incorporate}), referred as \textit{Src-Tgt} dual context. 

Another intuitive approach, as evident from Figure~\ref{fig:two-level}, is to separately model English and Foreign sentences using two separate context vectors $\mathbf{o}_{en,m}$ and $\mathbf{o}_{fr,m}$, where each is constructed from a mixture of the original source or target translations, is language-specific and possibly contain less noise. We refer to this as the \textit{Src-Tgt-Mix} dual context. Suppose $\mathbf{R}_{en,m}$ contains the mixed source/target representations for English (the dimensions for source representations have been reduced to \textit{H}) and $\mathbf{R}_{fr,m}$ contains the same for Foreign. Then,

\vspace{-2mm}
{\small
\begin{eqnarray*}
\mathbf{p}_{en,m}&=&\softmax(\mathbf{R}_{en,m}^T \times (\mathbf{W}_{td,en}\times \mathbf{h}_i+\mathbf{b}_{td,en}))\\		\nonumber
\mathbf{p}_{fr,m}&=&\softmax(\mathbf{R}_{fr,m}^T \times \tanh(\mathbf{W}_{tt,en}\times \mathbf{h}_i +\mathbf{b}_{tt,en}))\\		\nonumber
\mathbf{o}_{en,m}&=&\tanh(\mathbf{W}_{tr,en}\times(\mathbf{R}_{en,m}\times \mathbf{p}_{en,m})+\mathbf{b}_{tr,en})\\		\nonumber
\mathbf{o}_{fr,m}&=&\mathbf{R}_{fr,m}\times \mathbf{p}_{fr,m}		\nonumber
\end{eqnarray*}
}\normalsize
where $\mathbf{W}_{td,en}$, $\mathbf{W}_{tr,en}$ and $\mathbf{W}_{tt,en}$ are for dimensionality reduction, changing the language space and both, respectively.

\subsection{Incorporating Context into Base Model}\label{sec:incorporate}
The final representations $\mathbf{o}_{src}$ and $\mathbf{o}_{tgt}$ or $\mathbf{o}_{en,m}$ and $\mathbf{o}_{fr,m}$, can be incorporated together or individually in the base model by:

\begin{itemize}
\item \textbf{InitDec} Using a non-linear transformation to initialise the decoder, similar to \newcite{Wang:17}: $\mathbf{s}_{i,0} = \tanh(\mathbf{V}\times \mathbf{o}_{i}+\mathbf{b}_s)$, where $i$ is the sentence index in current turn $2k+1$, \{$\mathbf{V}$, $\mathbf{b}_s$\} are encoder-decoder specific parameters and $\mathbf{o}_{i}$ is either a single context vector or a concatenation (transformed) of the two.
\item \textbf{AddDec} As an auxiliary input to the decoder (similar to \newcite{Jean:17,Wang:17,Maruf:18}):
\vspace{-2mm}
\begin{eqnarray}
\mathbf{s}_{i,n} = \tanh({\vW}_{s}\cdot \mathbf{s}_{i,n-1}+{\vW}_{sn}\cdot \vE_T[y_{i,n}]+ \nonumber \\ 
{\vW}_{sc}\cdot {\vc}_{i,n}+ {\vW}_{ss}\cdot  \mathbf{o}_{i,src} +{\vW}_{st}\cdot  \mathbf{o}_{i,tgt} ) \nonumber
\end{eqnarray}
\item \textbf{InitDec+AddDec} Combination of previous two approaches.
\end{itemize}

\setlength{\tabcolsep}{0.5pt}
\begin{table*}[t!]
\centering
{\small
\begin{tabular}{l||c c c |c c c | c c c|| c c c}
& \multicolumn{9}{c}{\textbf{\small{Europarl}}} & \multicolumn{3}{||c}{\textbf{\small{Subtitles}}}\\
\cline{2-13}
& \multicolumn{3}{c}{\textbf{\small{En-Fr}}} & \multicolumn{3}{c}{\textbf{\small{En-Et}}} & \multicolumn{3}{c}{\textbf{\small{En-De}}} & \multicolumn{3}{||c}{\textbf{\small{En-Ru}}} \\
\cline{2-13}
& {\small{Overall}} & {\small{En$\rightarrow$Fr}} & {\small{Fr$\rightarrow$En}} & {\small{Overall}} & {\small{En$\rightarrow$Et}} & {\small{Et$\rightarrow$En}} & {\small{Overall}} & {\small{En$\rightarrow$De}} & {\small{De$\rightarrow$En}} & {\small{Overall}} & {\small{En$\rightarrow$Ru}} & {\small{Ru$\rightarrow$En}}\\
\hline
\hline
\small{\textit{Base Model}} & 37.36 & 38.13 & 36.03 & 20.68 & 18.64 & 26.65 & 24.74 & 21.80 & 27.74 & 19.05 & 14.90 & 23.04 \\
\hline
\hline
\noalign{\vskip 2mm}
\multicolumn{10}{l}{{\small{+\textit{Source Context as Lang-Specific Attention via}}}} & & &\\
\ \ \ \ \small{InitDec} & 38.40$^{\dagger}$ & 39.19$^{\dagger}$ & 36.86$^{\dagger}$ & \textbf{21.79}$^{\dagger}$ & 19.54$^{\dagger}$ & \textbf{28.33}$^{\dagger}$ & \textbf{26.34}$^{\dagger}$ & \textbf{23.31}$^{\dagger}$ & 29.39$^{\dagger}$ & 18.88 & 14.89 & 22.56\\
\ \ \ \ \small{AddDec} & 38.50$^{\dagger}$ & \textbf{39.35}$^{\dagger}$ & 36.98$^{\dagger}$ & 21.65$^{\dagger}$ & \textbf{19.66}$^{\dagger}$ & 27.48$^{\dagger}$ & 26.30$^{\dagger}$ & 23.09$^{\dagger}$ & \textbf{29.52}$^{\dagger}$ & 19.34 & 15.16 & 23.12\\
\ \ \ \ \small{InitDec+AddDec} & \textbf{38.55}$^{\dagger}$ & 39.34$^{\dagger}$ & \textbf{37.14}$^{\dagger}$ & 21.49$^{\dagger}$ & 19.43$^{\dagger}$ & 27.55$^{\dagger}$ & 26.25$^{\dagger}$ & 23.18$^{\dagger}$ & 29.30$^{\dagger}$ & \textbf{19.35} & \textbf{15.16} & \textbf{23.14}\\
\hline
\noalign{\vskip 2mm}
\multicolumn{10}{l}{{\small{+\textit{Source Context via}}}} & & &\\
\ \ \ \ \small{Direct Tranformation} & 38.35$^{\dagger}$ & 39.13$^{\dagger}$ & 36.96$^{\dagger}$ & 21.75$^{\dagger}$ & \textbf{19.59}$^{\dagger}$ & 28.07$^{\dagger}$ & 26.29$^{\dagger}$ & 23.34$^{\dagger}$ & 29.22$^{\dagger}$ & 19.09 & 14.89 & 22.76\\
\ \ \ \ \small{Hierarchical Gating} & 38.33$^{\dagger}$ & 39.14$^{\dagger}$ & 36.89$^{\dagger}$ & 21.62$^{\dagger}$ & 19.55$^{\dagger}$ & 27.64$^{\dagger}$ & 26.31$^{\dagger}$ & 23.17$^{\dagger}$ & 29.45$^{\dagger}$ & 19.20 & 15.10 & 22.73\\
\ \ \ \ \small{Lang-Specific Attention} & 38.40$^{\dagger}$ & 39.19$^{\dagger}$ & 36.86$^{\dagger}$ & 21.79$^{\dagger}$ & 19.54$^{\dagger}$ & 28.33$^{\dagger}$ & 26.34$^{\dagger}$ & 23.31$^{\dagger}$ & 29.39$^{\dagger}$ & \textbf{19.35} & \textbf{15.16} & \textbf{23.14}\\
\ \ \ \ \small{Combined Attention} & \textbf{38.50}$^{\dagger}$ & \textbf{39.36}$^{\dagger}$ & 36.94$^{\dagger}$ & 21.66$^{\dagger}$ & 19.52$^{\dagger}$ & 27.90$^{\dagger}$ & 26.38$^{\dagger}$ & 23.31$^{\dagger}$ & 29.44$^{\dagger}$ & 18.96 & 14.82 & 22.92\\
\ \ \ \ \small{Lang-Specific S-Attention} & 38.46$^{\dagger}$ & 39.24$^{\dagger}$ & \textbf{37.06}$^{\dagger}$ & \textbf{21.84}$^{\dagger}$ & 19.58$^{\dagger}$ & \textbf{28.43}$^{\dagger}$ & \textbf{26.49}$^{\dagger}$ & \textbf{23.49}$^{\dagger}$ & \textbf{29.49}$^{\dagger}$ & 19.09 & 14.59 & 22.98\\
\hline
\noalign{\vskip 2mm}
\multicolumn{10}{l}{{\small{+\textit{Lang-Specific S-Attention using}}}} & & &\\
\ \ \ \ \small{Source Context} & 38.46$^{\dagger}$ & 39.24$^{\dagger}$ & 37.06$^{\dagger}$ & \textbf{21.84}$^{\dagger}$ & 19.58$^{\dagger}$ & \textbf{28.43}$^{\dagger}$ & \textbf{26.49}$^{\dagger}$ & \textbf{23.49}$^{\dagger}$ & 29.49$^{\dagger}$ & 19.09 & 14.59 & 22.98\\
\ \ \ \ \small{Target Context} & 38.76$^{\dagger}$ & \textbf{39.57}$^{\dagger}$ & 37.35$^{\dagger}$ & 21.77$^{\dagger}$ & \textbf{19.68}$^{\dagger}$ & 27.86$^{\dagger}$ & 26.21$^{\dagger}$ & 23.16$^{\dagger}$ & 29.26$^{\dagger}$ & 19.23 & 14.77 & \textbf{23.23}\\
\ \ \ \ \small{Dual Context Src-Tgt} & \textbf{38.80}$^{\dagger}$ & 39.51$^{\dagger}$ & \textbf{37.50}$^{\dagger}$ & 21.74$^{\dagger}$ & 19.60$^{\dagger}$ & 27.98$^{\dagger}$ & 26.39$^{\dagger}$ & 23.28$^{\dagger}$ & \textbf{29.50}$^{\dagger}$ & 18.89 & 14.52 & 23.06\\
\ \ \ \ \small{Dual Context Src-Tgt-Mix} & 38.76$^{\dagger}$ & 39.52$^{\dagger}$ & 37.43$^{\dagger}$ & 21.68$^{\dagger}$ & 19.63$^{\dagger}$ & 27.71$^{\dagger}$ & 26.37$^{\dagger}$ & 23.26$^{\dagger}$ & 29.48$^{\dagger}$ & \textbf{19.26} & \textbf{14.86} & 23.01\\
\hline
\end{tabular}
}
\caption{BLEU scores for the bilingual test sets. Here all contexts are incorporated as InitDec for Europarl and InitDec+AddDec for Subtitles unless otherwise specified. \textbf{bold}: Best performance, $\dagger$: Statistically significantly better than the base model, based on bootstrap resampling \cite{Clark:11} with \textit{p} $<$ 0.05.}
\label{table:allresults}
\end{table*}

\subsection{Training and Decoding}
The model parameters are trained end-to-end by maximising the sum of log-likelihood of the bilingual conversations in training set $\mathcal{D}$. For example, for a conversation having alternating turns of English and Foreign language, the log-likelihood is:

\vspace{-4mm}
{\small
\begin{eqnarray*}
\sum_{k=0}^{\frac{|T|}{2}-1}\Big({\sum_{i=1}^{|t_{2k+1}|}\log P_{\vtheta}(\vy_{i}|\vx_{i}, \mathbf{o}_{i}) + \sum_{j=1}^{|t_{2k+2}|}\log P_{\vtheta}(\vx_{j} |\vy_{j}, \mathbf{o}_{j})}\Big)
\end{eqnarray*}
}\normalsize
where $i$, $j$ denote sentences belonging to ${2k+1}^{th}$ or ${2k+2}^{th}$ turn; $\mathbf{o}_{(.)}$ is a representation of the conversation history, and $|T|$ is the total number of turns (assumed to be even here).

The best output sequence for a given input sequence for the $i^{th}$ sentence at test time, a.k.a. decoding, is produced by: $$\arg\max_{\vy_i} P_{\vtheta}(\vy_i |\vx_i, \mathbf{o}_{i})$$

\section{Experiments}
\paragraph{Implementation and Hyperparameters}

We implement our conversational Bi-MSMT model in C++ using the DyNet library \cite{dynet}. The base model is built using \texttt{mantis} \cite{Cohn:16} which is an implementation of the generic sentence-level NMT model using DyNet. 

The base model has single layer bidirectional GRUs in the encoder and 2-layer GRU in the decoder\footnote{We follow \newcite{Cohn:16} and \newcite{Britz:2017} in choosing hyperparameters for our model.}. The hidden dimensions and word embedding sizes are set to 256, and the alignment dimension (for the attention mechanism in the decoder) is set to 128. 

\paragraph{Models and Training}

We do a stage-wise training for the base model, i.e., we first train the English$\rightarrow$Foreign architecture and the Foreign$\rightarrow$English architecture, using the sentence-level parallel corpus. Both architectures have the same vocabulary\footnote{For each language-pair, we use BPE \cite{Sennrich:16} to obtain a joint vocabulary of size $\approx$30k.} but separate parameters to avoid biasing the embeddings towards the architecture trained last. The contextual model is pre-trained similar to training the base model. The best model is chosen based on minimum overall perplexity on the bilingual dev set. 

For the source context representations, we use the sentence representations generated by two sentence-level bidirectional RNNLMs (one each for English and Foreign) trained offline.
For the target sentence representations, 
we use the last hidden states of the decoder generated from the pre-trained base model\footnote{Even though the paramaters of the base model are updated, the target sentence representations are fixed throughout training. We experimented with a scheduled updating scheme in preliminary experiments but it did not yield significant improvement.}. 
At decoding time, however, we use the last hidden state of the decoder computed by our model (not the base) as the target sentence representations. 
Further training details are provided in Appendix~\ref{app:exp}.



\subsection{Results}

Firstly, we evaluate the three strategies for incorporating context: InitDec, AddDec, InitDec+AddDec, and report the results for source context using Language-Specific Attention in Table~\ref{table:allresults}. 
For the Europarl data, we see decent improvements with InitDec for En-Et (+1.11 BLEU) and En-De (+1.60 BLEU), and with InitDec+AddDec for En-Fr (+1.19 BLEU). We also observe that, for all language-pairs, both translation directions benefit from context, showing that our training methodology was indeed effective. On the other hand, for the Subtitles data, we see a maximum improvement of +0.30 BLEU for InitDec+AddDec
. We narrow down to three major reasons: (i) the data is noisier when compared to Europarl, (ii) the sentences are short and generic with only 1\% having more than 27 tokens, and finally (iii) the turns in OpenSubtitles2016 are short compared to those in Europarl (see Table~\ref{tab:data}), and we show later (Section~\ref{sec:analysis}) that the context from current turn is the most important.

The next set of experiments evaluates the five different approaches for computing the source-side context. It is evident from Table~\ref{table:allresults} that for English-Estonian and English-German, our model indeed benefits from using the fine-grained sentence-level information (Language-Specific Sentence-level Attention) as opposed to just the turn-level one. 

Finally, our results with source, target and dual contexts are reported. Interestingly, just using the source context is sufficient for English-Estonian and English-German. 
For English-French, on the other hand, we see significant improvements for the models using the target-side conversation history over using only the source-side. We attribute this to the base model being more efficient and able to generate better translations for En-Fr as it had been trained on a larger corpus as opposed to the other two language-pairs. Unlike Europarl, for Subtitles, we see improvements for our Src-Tgt-Mix dual context variant over the Src-Tgt one for En$\rightarrow$Ru, showing this to be an effective approach when the target representations are noisier.

To summarise, for majority of the cases our Language-Specific Sentence-level Attention is a winner or a close second. Using the Target Context is useful when the base model generates reasonable-quality translations; otherwise, using the Source Context should suffice. 

\paragraph{Local Source Context Model}
Most of the previous works for online context-based NMT consider only a single previous sentence as context \cite{Jean:17,Bawden:17,Voita:18}.  
Drawing inspiration from these works, we evaluate our model (trained with Language-Specific Sentence-Level Attention) on the same test set but using only the previous source sentence as context. This evaluation allows us to hypothesise how much of the gain can be attributed to the previous sentence. From Table~\ref{table:auxresults}, it can be seen that our model surpasses the local-context baseline for Europarl showing that the wider context is indeed beneficial if the turn lengths are longer. For En-Ru, it can be seen that using previous sentence is sufficient due to short turns (see Table~\ref{tab:data}).

\setlength{\tabcolsep}{2pt}
\begin{table}[t!]
\centering
{\small
\begin{tabular}{l|c c c |c }
& \multicolumn{3}{c}{\textbf{\small{Europarl}}} & \textbf{\small{Subtitles}}\\
\cline{2-5}
& {\textbf{\small{En-Fr}}} & {\textbf{\small{En-Et}}} & {\textbf{\small{En-De}}} & {\textbf{\small{En-Ru}}} \\
\hline
\textit{Prev Sent} & 38.15 & 21.70 & 26.09 & \textbf{19.13}\\
\hline
Our Model & \textbf{38.46}$^{\dagger}$ & \textbf{21.84} & \textbf{26.49}$^{\dagger}$ & 19.09\\
\hline
\end{tabular}
}
\caption{BLEU scores for the bilingual test sets. \textbf{bold}: Best performance, $\dagger$: Statistically significantly better than the contextual baseline.}
\label{table:auxresults}
\end{table}

\setlength{\tabcolsep}{2pt}
\begin{table}[t!]
\centering
{\small
\begin{tabular}{l|c }
\textbf{\small{Type of Context}} & \textbf{\small{BLEU}}\\
\hline
\hline
\small{No context (Base Model)} & 24.74\\
\hline
\small{Current Turn} & 26.39\\
\small{Current Language from Previous Turns} & 26.21\\
\small{Other Language from Previous Turns} & 26.32\\
\small{Complete Context} & \textbf{26.49}\\
\hline
\end{tabular}
}
\caption{BLEU scores for En-De bilingual test set.}
\label{table:ablation}
\end{table}

\pgfplotstableread[row sep=\\,col sep=&]{
    size & base & context\\
    I & 24.74 & 26.34\\
    II & 28.23 & 29.35\\
    III & 29.75 & 31.23\\
    }\mydata

\begin{figure}[t!]
\begin{tikzpicture}
    \begin{axis}[
            ybar,
	enlarge x limits=0.2,
            x=2cm,
            bar width=0.6cm,
            height=.63\linewidth,
            legend style={at={(0.5,1)},
                anchor=north,legend columns=-1},
            symbolic x coords={I, II, III},
            xtick=data,
            nodes near coords,
 	 every node near coord/.append style={font=\scriptsize},
            nodes near coords align={vertical},
            ymin=24,ymax=34,
            ylabel={BLEU},
        ]
        \addplot table[x=size,y=base]{\mydata};
        \addplot table[x=size,y=context]{\mydata};
        \legend{\small{Base MT}, \small{BaseMT+SrcContext}}
    \end{axis}
\end{tikzpicture}
\caption{BLEU scores on En-De test set while training (I) smaller base model with smaller corpus (previous experiment), (II) smaller base model with larger corpus, and (III) a larger base model with larger corpus.}
\label{fig:compare}
\end{figure}
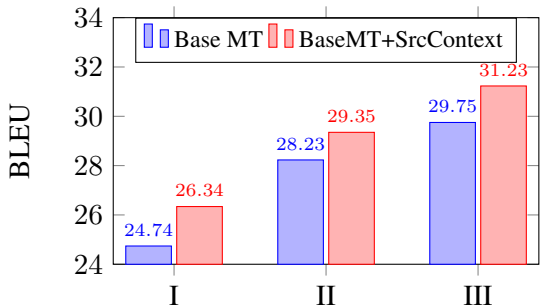

\subsection{Analysis}\label{sec:analysis}
\paragraph{Ablation Study}
We conduct an ablation study to validate our hypothesis of using the complete context versus using only one of the three types of contexts in a bilingual multi-speaker conversation: (i) current turn, (ii) previous turns in current language, and (iii) previous turns in the other language.
The results for En-De are reported in Table~\ref{table:ablation}. We see decrease in BLEU for all types of contexts with significant decrease when considering only current language from previous turns.
The results show that the current turn has the most influence on translating a sentence, and we conclude that since our model is able to capture the complete context, it is generalisable to any conversational scenario.

\paragraph{Training base model with more data}

To analyse if the context is beneficial even when using more data, we perform an experiment for English-German 
 where we train the base model with additional sentence-pairs from the full WMT'14 corpus\footnote{https://nlp.stanford.edu/projects/nmt/} (excluding our dev/test sets and filtering sentences with more than 100 tokens). For training the contextual model, we still use the bilingual multi-speaker corpus. We observe a significant improvement of +1.12 for the context-based model (Figure~\ref{fig:compare} II), showing the significance of conversation history in this experiment condition.\footnote{It should be noted that the BLEU score for the base model trained with WMT does not match the published results exactly as the test set contains both English and German sentences. It does, however, fall between the scores usually obtained on WMT'14 for En$\rightarrow$De and De$\rightarrow$En.}

We perform another experiment where we use a larger base model, having almost double the number of parameters than our previous base model (hidden units and word embedding sizes set to 512, and alignment dimension set to 256), to test if the model parameters are being overestimated due to the additional context. We use the same WMT'14 corpus to train the base model and achieve significant improvement of +1.48 BLEU for our context-based model over the larger baseline (Figure~\ref{fig:compare} III).

\setlength{\tabcolsep}{0.5pt}
\begin{table}[t!]
\centering
{\small
\begin{tabular}{l|l}
\hline
\hline
En$\rightarrow$Fr & les; par; est; a; dans; le; en; j'; un; afin; question;\\
& entre; qu'; \^etre; ces; {\'e}galement; y; depuis; c'; ou\\
\hline
Fr$\rightarrow$En & this; of; we; issue; europe; by; up; make; united;\\
& does; what; regard; s; must; however; such; whose; \\
& share; like; been\\
\hline
\hline
En$\rightarrow$Et & eest; vahel; {\"u}le; nimel; ja; aastal; aasta; neid; ainult\\
& seep{\"a}rast; nagu; kes; komisjoni; tehtud; k{\"u}simuses; \\
& sisser{\"a}nde; liikmesriigi; mulla; liibanoni; dawit\\
\hline
Et$\rightarrow$En & for; this; of; is; political; important; culture; also; as;\\
& order; are; each; their; only; gender; were; its; \\
& economy; one; market\\
\hline
\hline
En$\rightarrow$De & da{\ss}; auf; und; werden; nicht; m{\"u}ssen; aus; mehr;\\
& k{\"o}nnen; einem; rates; eines; insbesondere; wurden; \\
& habe; mitgliedstaaten; ist; sondern; europa; \\
& gemeinsamen\\
\hline
De$\rightarrow$En & that; its; say; must; some; therefore; more; countries; \\
& an; favour; public; will; without; particularly; \\
& hankiss; much; increase; eu; them; parliamentary\\
\hline
\hline
\end{tabular}
}
\caption{Most frequent tokens correctly generated by our model when compared to the base model.}
\label{tab:freq-words}
\end{table}

\begin{figure}[t!]
 \centering
  \includegraphics[width=0.26\textwidth]{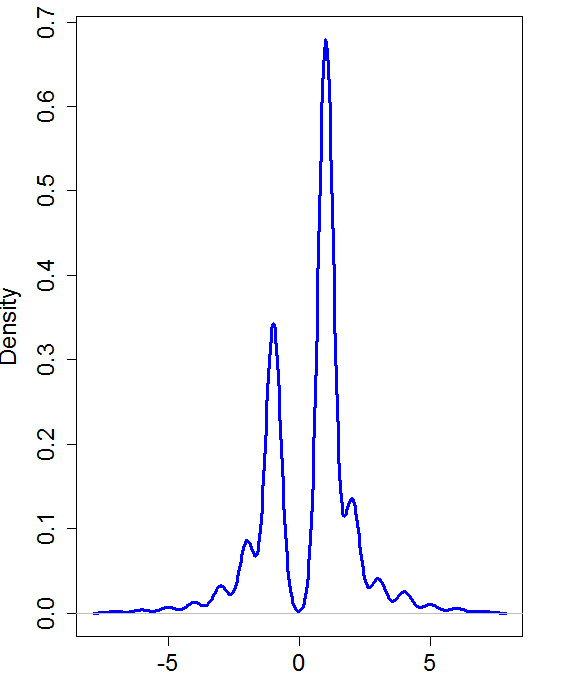}
 \caption{Density of token counts for En$\rightarrow$Fr illustrating where our model is better (+ve x-axis) and where the base model is better (-ve x-axis).}
\label{fig:density}
\end{figure}

\paragraph{How is the context helping?}

\setlength{\tabcolsep}{0.4pt}
\begin{table*}[t]
\centering
{\small
\begin{tabular}{|l|l|}
\hline
\textit{Context} & nous sommes {\'e}galement favorables au principe d'un syst{\`e}me de collecte des miles commun pour le parlement \\
& europ{\'e}en, pour que celui-ci puisse b{\'e}n{\'e}ficier de billets d'avion moins chers, m{\^e}me si nous voyons difficilement \\
& comment ce syst{\`e}me pourrait {\^e}tre d{\'e}ploy{\'e} en pratique.\\
& enfin, nous ne sommes pas oppos{\'e}s {\`a} l'attribution de prix culturels par le parlement europ{\'e}en.\\
\hline
\textit{Source} & {\color{blue}n{\'e}anmoins}, nous sommes particuli{\`e}rement critiques {\`a} l'{\'e}gard du prix pour le journalisme du parlement europ{\'e}en\\
& et nous ne pensons pas que celui-ci puisse d{\'e}cerner des prix aux journalistes ayant pour mission de soumettre le \\
& parlement europ{\'e}en {\`a} un regard critique. \\
\textit{Target} & {\color{blue}however}, we are highly critical of parliament's prize for journalism, and do not believe that it is appropriate for \\
& parliament to award prizes to journalists whose task it is to critically examine the european parliament.\\
\hline
\textit{Base Model} & {\color{red}nevertheless}, we are particularly critical of the price for the european union's european alism and we do not \\
& believe that it would be able to make a price to the journalists who have been made available to the european \\
& parliament to a critical view.\\
\textit{Our Model} & {\color{blue}however}, we are particularly critical of the price for the european union's democratic alism and we do not believe \\
& that it can give rise to the prices for journalists who have been tabled to submit the european parliament to a \\
& critical view. \\
\hline                  
\end{tabular}
}
\caption{Example En-Fr sentence translation showing how the context helps our model in generating the appropriate discourse connective.}
\label{table:exhowever}
\end{table*}

\setlength{\tabcolsep}{0.4pt}
\begin{table*}[t]
\centering
{\small
\begin{tabular}{|l|l|}
\hline
\textit{Context} & oleks hea, kui {\color{blue}reitinguagentuurid} vastutaksid tulevikus enda tegevuse eest rohkem.\\
& ...\\
& kirjalikult. - (it) kiites heaks wolf klinzi raporti, mille eesm{\"a}rk on {\color{blue}reitinguagentuuride} t{\~o}hus reguleerimine,\\
& v{\~o}tab parlament j{\"a}rjekordse sammu finantsturgude suurema l{\"a}bipaistvuse suunas.\\
& ...\\
& mul oli selle dokumendi {\"u}le hea meel, sest {\color{blue}krediidireitingute valdkonnal} on palju probleeme, millest k{\~o}ige \\
& suuremad on oligopolidele t{\"u}pilised struktuurid ning konkurentsi, vastutuse ja l{\"a}bipaistvuse puudumine. \\
\hline
\textit{Source} & selles suhtes tuleb r{\~o}hutada {\color{blue}nende} tegevuse suuremal {\"a}bipaistvuse p{\~o}hirolli. \\
\textit{Target} & in this respect, it is necessary to highlight the central role of increased transparency in {\color{blue}their} activities. \\
\hline
\textit{Base Model} & in this regard it must be emphasised in the major role of transparency in which {\color{red}these} activities are to be \\
& strengthened. \\
\textit{Our Model} & in this regard, it must be stressed in the key role of greater transparency in {\color{blue}their} activities. \\
\hline                  
\end{tabular}
}
\caption{Example En-Et translation showing how the wide-range context helps in generating the correct pronoun. The antecedent and correct pronoun are highlighted in blue.}
\label{table:expronoun}
\end{table*}

The underlying hypothesis for this work is that discourse phenomenon in a conversation may depend on long-range dependency and these may be ignored by the sentence-based NMT models. To analyse if our contextual model is able to accurately translate such linguistic phenomenon, 
we come up with our own evaluation procedure. 
We aggregate the tokens correctly generated by our model and those correctly generated by the baseline over the entire test set. We then take the difference of these counts and sort them\footnote{We do not normalise the counts with the background frequency as it favours rare words. Thus, obscuring the main reasons of improving the BLEU score.}. Table~\ref{tab:freq-words} reports the top 20 tokens where our model is better than the baseline for the Europarl dataset. 
Figure~\ref{fig:density} gives the density of counts obtained using our evaluation for En$\rightarrow$Fr\footnote{Outliers and tokens with equal counts for our model and the baseline were removed.}. Positive counts correspond to correct translations by our model while the negative counts correspond to where the base model was better. It can be seen that for majority of cases our model supersedes the base model. We observed a similar trend for other translation directions. 
In general, the correctly generated tokens by our model include pronouns (that, this, its, their, them), discourse connectives (e.g., `however', `therefore', `also') and prepositions (of, for, by). 

\begin{figure}[t!]
 \centering
  \includegraphics[width=0.43\textwidth, height=0.33\textwidth]{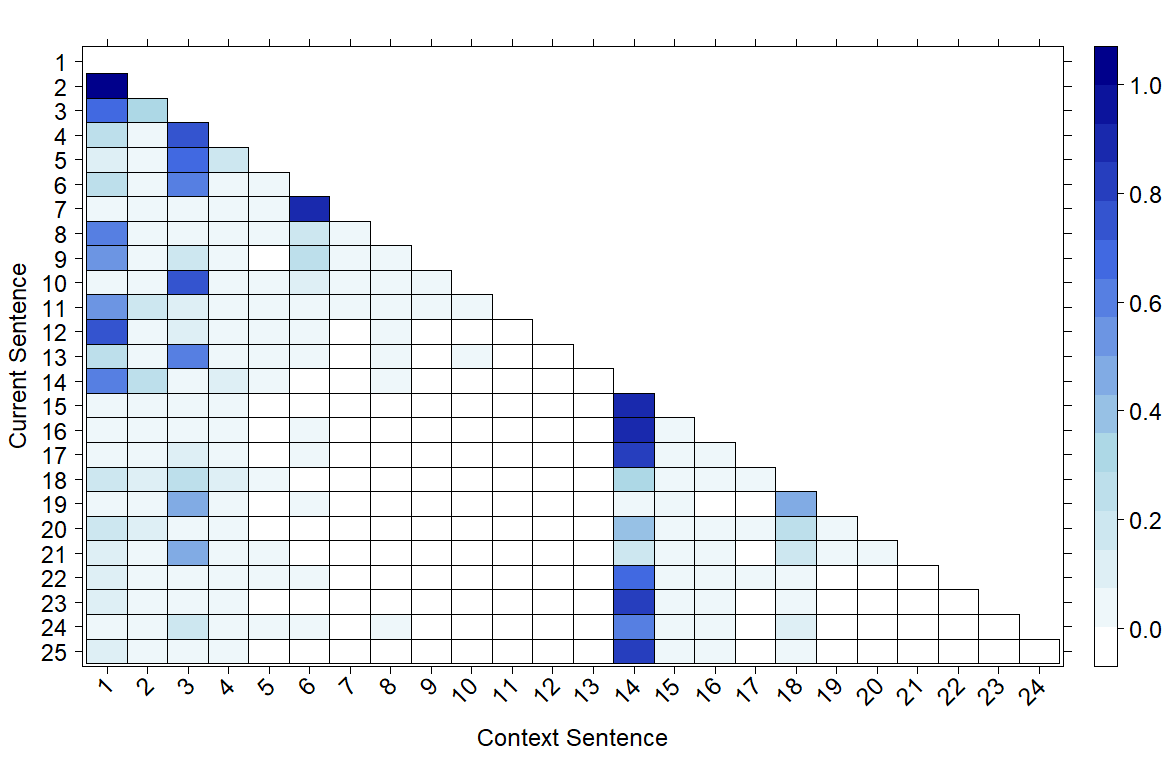}
 \caption{Attention map when translating a conversation from the Et-En test set.}
\label{fig:heatmap}
\end{figure}

Table~\ref{table:exhowever} reports an example where our model is able to generate the correct discourse connective `\textit{however}' using the context. If we look at the context of the source sentence in French, we come to the conclusion that `however' is indeed a perfect fit in this case, whereas the base model is at a disadvantage and completely changes the underlying meaning of the sentence by generating the inappropriate connective `nevertheless'. 

Table~\ref{table:expronoun} gives an instance where our model is able to generate the correct pronoun `\textit{their}'. It should be noted that in this case, the current source sentence does not contain the antecedent and thus the context-free baseline is unable to generate the appropriate pronoun. On the other hand, our contextual model is able to do so by giving the highest attention weights to sentences containing the antecedent (observed from the attention map in Figure~\ref{fig:heatmap})\footnote{For this particular conversation, all previous turns were in Estonian.}. Figure \ref{fig:heatmap} also shows that for translating majority of the sentences, the model attends to wide-range context rather than just the previous sentence, hence strengthening the premise of using the complete context. 

\section{Conclusion}
This work investigates the challenges associated with translating multilingual multi-speaker conversations by exploring a simpler task referred to as Bilingual Multi-Speaker Conversation MT. We process Europarl v7 and OpenSubtitles2016 to obtain an introductory dataset for this task. Compared to models developed for similar tasks, our work is different in two aspects: (i) the history captured by our model contains multiple languages, and (ii) our model captures `global' history as opposed to `local' history captured in most previous works. Our experiments demonstrate the significance of leveraging the bilingual conversation history in such scenarios. Furthermore, the analysis shows that using wide-range context, our model generates appropriate pronouns and discourse connectives in some cases. We hope this work to be a first step towards translating multilingual multi-speaker conversations. Future work on this task may include optimising the base translation model and approaches that condition on specific discourse information in the conversation history.   

\section*{Acknowledgments}
We thank the reviewers for their useful feedback that helped us in improving our work. We also thank Pierre Lison for providing us with the speaker-annotated OpenSubtitles2016 dataset. This work is supported by the Multi-modal Australian ScienceS Imaging and Visualisation Environment (MASSIVE) (\url{www.massive.org.au}), the European Research Council (ERC StG DeepSPIN 758969), the Funda\c{c}\~ao para a Ci\^encia e Tecnologia through contract UID/EEA/50008/2013, a Google Faculty Research Award to GH and the Australian Research Council through DP160102686.

\bibliography{emnlp2018}

\begin{thebibliography}{28}
\expandafter\ifx\csname natexlab\endcsname\relax\def\natexlab#1{#1}\fi

\bibitem[{Bahdanau et~al.(2015)Bahdanau, Cho, and Bengio}]{Bahdanau:15}
Dzmitry Bahdanau, Kyunghyun Cho, and Yoshua Bengio. 2015.
\newblock Neural machine translation by jointly learning to align and
  translate.
\newblock In \emph{Proceedings of the International Conference on Learning
  Representations}.

\bibitem[{Bawden et~al.(2017)Bawden, Sennrich, Birch, and Haddow}]{Bawden:17}
Rachel Bawden, Rico Sennrich, Alexandra Birch, and Barry Haddow. 2017.
\newblock Evaluating discourse phenomena in neural machine translation.
\newblock In \emph{Proceedings of NAACL--HLT 2018}.

\bibitem[{Britz et~al.(2017)Britz, Goldie, Luong, and Le}]{Britz:2017}
Denny Britz, Anna Goldie, Minh-Thang Luong, and Quoc Le. 2017.
\newblock Massive exploration of neural machine translation architectures.
\newblock In \emph{Proceedings of the Conference on Empirical Methods in
  Natural Language Processing}. Association for Computational Linguistics.

\bibitem[{Cho et~al.(2014)Cho, {van Merrienboer}, Bahdanau, and
  Bengio}]{Cho:14}
Kyunghyun Cho, B~{van Merrienboer}, Dzmitry Bahdanau, and Yoshua Bengio. 2014.
\newblock On the properties of neural machine translation: Encoder-decoder
  approaches.
\newblock In \emph{Eighth Workshop on Syntax, Semantics and Structure in
  Statistical Translation (SSST-8)}.

\bibitem[{Clark et~al.(2011)Clark, Dyer, Lavie, and Smith}]{Clark:11}
Jonathan~H. Clark, Chris Dyer, Alon Lavie, and Noah~A. Smith. 2011.
\newblock Better hypothesis testing for statistical machine translation:
  Controlling for optimizer instability.
\newblock In \emph{Proceedings of the 49th Annual Meeting of the Association
  for Computational Linguistics: Human Language Technologies (Short Papers)},
  pages 176--181. Association for Computational Linguistics.

\bibitem[{Cohn et~al.(2016)Cohn, Hoang, Vymolova, Yao, Dyer, and
  Haffari}]{Cohn:16}
Trevor Cohn, Cong Duy~Vu Hoang, Ekaterina Vymolova, Kaisheng Yao, Chris Dyer,
  and Gholamreza Haffari. 2016.
\newblock Incorporating structural alignment biases into an attentional neural
  translation model.
\newblock In \emph{Proceedings of the North American Chapter of the Association
  for Computational Linguistics: Human Language Technologies}, pages 876--885.
  Association for Computational Linguistics.

\bibitem[{Garcia et~al.(2017)Garcia, Creus, Espa{\~n}a-Bonet, and
  M{\`a}rquez}]{Garcia:17}
Eva~Mart{\'i}nez Garcia, Carles Creus, Cristina Espa{\~n}a-Bonet, and Llu{\'i}s
  M{\`a}rquez. 2017.
\newblock Using word embeddings to enforce document-level lexical consistency
  in machine translation.
\newblock \emph{The Prague Bulletin of Mathematical Linguistics}, 108:85--96.

\bibitem[{Garcia et~al.(2014)Garcia, Espa{\~n}a-Bonet, and
  M{\`a}rquez}]{Garcia:14}
Eva~Mart{\'i}nez Garcia, Cristina Espa{\~n}a-Bonet, and Llu{\'i}s M{\`a}rquez.
  2014.
\newblock Document-level machine translation as a re-translation process.
\newblock \emph{Procesamiento del Lenguaje Natural}, 53:103--110.

\bibitem[{Gong et~al.(2011)Gong, Zhang, and Zhou}]{Gong:11}
Zhengxian Gong, Min Zhang, and Guodong Zhou. 2011.
\newblock Cache-based document-level statistical machine translation.
\newblock In \emph{Proceedings of the Conference on Empirical Methods in
  Natural Language Processing}, pages 909--919. Association for Computational
  Linguistics.

\bibitem[{Hardmeier and Federico(2010)}]{Hardmeier:10}
Christian Hardmeier and Marcello Federico. 2010.
\newblock Modelling pronominal anaphora in statistical machine translation.
\newblock In \emph{International Workshop on Spoken Language Translation},
  pages 283--289.

\bibitem[{Hardmeier et~al.(2012)Hardmeier, Nivre, and Tiedemann}]{Hardmeier:12}
Christian Hardmeier, Joakim Nivre, and J\"{o}rg Tiedemann. 2012.
\newblock Document-wide decoding for phrase-based statistical machine
  translation.
\newblock In \emph{Proceedings of the 2012 Joint Conference on Empirical
  Methods in Natural Language Processing and Computational Natural Language
  Learning}, pages 1179--1190. Association for Computational Linguistics.

\bibitem[{Hoang et~al.(2016)Hoang, Cohn, and Haffari}]{Hoang:16}
Cong Duy~Vu Hoang, Trevor Cohn, and Gholamreza Haffari. 2016.
\newblock Incorporating side information into recurrent neural network language
  models.
\newblock In \emph{Proceedings of NAACL--HLT 2016}, pages 1250--1255.

\bibitem[{Jean et~al.(2017)Jean, Lauly, Firat, and Cho}]{Jean:17}
Sebastien Jean, Stanislas Lauly, Orhan Firat, and Kyunghyun Cho. 2017.
\newblock Does neural machine translation benefit from larger context?
\newblock In \emph{arXiv:1704.05135}.

\bibitem[{Ji and Bilmes(2004)}]{Ji:04}
Gang Ji and Jeff Bilmes. 2004.
\newblock Multi-speaker language modeling.
\newblock In \emph{Proceedings of HLT--NAACL 2004}.

\bibitem[{Ji et~al.(2015)Ji, Cohn, Kong, Dyer, and Eisenstein}]{Ji:15}
Yangfeng Ji, Trevor Cohn, Lingpeng Kong, Chris Dyer, and Jacob Eisenstein.
  2015.
\newblock Document context language models.
\newblock In \emph{Workshop track - ICLR 2016}.

\bibitem[{Koehn(2005)}]{Koehn:05}
Philipp Koehn. 2005.
\newblock Europarl: A parallel corpus for statistical machine translation.
\newblock In \emph{Conference Proceedings: the 10th Machine Translation
  Summit}, pages 79--86. AAMT.

\bibitem[{Kuang et~al.(2018)Kuang, Xiong, Luo, and Zhou}]{Kuang:18}
Shaohui Kuang, Deyi Xiong, Weihua Luo, and Guodong Zhou. 2018.
\newblock Modeling coherence for neural machine translation with dynamic and
  topic caches.
\newblock \emph{COLING 2018}.

\bibitem[{Lison and Meena(2016)}]{LisonMeena:2016}
Pierre Lison and Raveesh Meena. 2016.
\newblock Automatic turn segmentation of movie \& tv subtitles.
\newblock In \emph{Proceedings of the 2016 Spoken Language Technology
  Workshop}, pages 245--252, San Diego, CA, USA. IEEE.

\bibitem[{Lison and Tiedemann(2016)}]{Lison:16}
Pierre Lison and J{\"o}rg Tiedemann. 2016.
\newblock Open{S}ubtitles2016: Extracting large parallel corpora from {M}ovie
  and {TV} subtitles.
\newblock In \emph{Proceedings of the 10$^{th}$ International Conference on
  Language Resources and Evaluation (LREC'16)}, pages 923--929.

\bibitem[{Maruf and Haffari(2018)}]{Maruf:18}
Sameen Maruf and Gholamreza Haffari. 2018.
\newblock Document context neural machine translation with memory networks.
\newblock In \emph{Proceedings of the 56th Annual Meeting of the Association
  for Computational Linguistics}. Association for Computational Linguistics.

\bibitem[{Neubig et~al.(2017)Neubig, Dyer, Goldberg, Matthews, Ammar,
  Anastasopoulos, Ballesteros, Chiang, Clothiaux, Cohn, Duh, Faruqui, Gan,
  Garrette, Ji, Kong, Kuncoro, Kumar, Malaviya, Michel, Oda, Richardson,
  Saphra, Swayamdipta, and Yin}]{dynet}
Graham Neubig, Chris Dyer, Yoav Goldberg, Austin Matthews, Waleed Ammar,
  Antonios Anastasopoulos, Miguel Ballesteros, David Chiang, Daniel Clothiaux,
  Trevor Cohn, Kevin Duh, Manaal Faruqui, Cynthia Gan, Dan Garrette, Yangfeng
  Ji, Lingpeng Kong, Adhiguna Kuncoro, Gaurav Kumar, Chaitanya Malaviya, Paul
  Michel, Yusuke Oda, Matthew Richardson, Naomi Saphra, Swabha Swayamdipta, and
  Pengcheng Yin. 2017.
\newblock Dynet: The dynamic neural network toolkit.
\newblock \emph{arXiv preprint arXiv:1701.03980}.

\bibitem[{Sennrich et~al.(2016)Sennrich, Haddow, and Birch}]{Sennrich:16}
Rico Sennrich, Barry Haddow, and Alexandra Birch. 2016.
\newblock Neural machine translation of rare words with subword units.
\newblock In \emph{Proceedings of the 54$^{th}$ Annual Meeting of the
  Association for Computational Linguistics}, pages 1715--1725.

\bibitem[{Tran et~al.(2016)Tran, Zuckerman, and Haffari}]{Tran:16}
Quan~Hung Tran, Ingrid Zuckerman, and Gholamreza Haffari. 2016.
\newblock Inter-document contextual language model.
\newblock In \emph{Proceedings of NAACL--HLT 2016}, pages 762--766.

\bibitem[{Tu et~al.(2017)Tu, Liu, Shi, and Zhang}]{Tu:2017}
Zhaopeng Tu, Yang Liu, Shuming Shi, and Tong Zhang. 2017.
\newblock Learning to remember translation history with a continuous cache.

\bibitem[{Voita et~al.(2018)Voita, Serdyukov, Sennrich, and Titov}]{Voita:18}
Elena Voita, Pavel Serdyukov, Rico Sennrich, and Ivan Titov. 2018.
\newblock Context-aware neural machine translation learns anaphora resolution.
\newblock In \emph{Proceedings of the 56th Annual Meeting of the Association
  for Computational Linguistics}. Association for Computational Linguistics.

\bibitem[{Wang et~al.(2017)Wang, Tu, Way, and Liu}]{Wang:17}
Longyue Wang, Zhaopeng Tu, Andy Way, and Qun Liu. 2017.
\newblock Exploiting cross-sentence context for neural machine translation.
\newblock In \emph{Proceedings of the Conference on Empirical Methods in
  Natural Language Processing}, pages 2816--2821. Association for Computational
  Linguistics.

\bibitem[{Wang et~al.(2016)Wang, Zhang, Tu, Way, and Liu}]{Wang:2016}
Longyue Wang, Xiaojun Zhang, Zhaopeng Tu, Andy Way, and Qun Liu. 2016.
\newblock Automatic construction of discourse corpora for dialogue translation.
\newblock In \emph{Proceedings of the Tenth International Conference on
  Language Resources and Evaluation ({LREC} 2016)}, Paris, France. European
  Language Resources Association (ELRA).

\bibitem[{van~der Wees et~al.(2016)van~der Wees, Bisazza, and Monz}]{Wees:2016}
Marlies van~der Wees, Arianna Bisazza, and Christof Monz. 2016.
\newblock Measuring the effect of conversational aspects on machine translation
  quality.
\newblock In \emph{Proceedings of {COLING} 2016}, pages 2571--2581.

\end{thebibliography}
\bibliographystyle{acl_natbib_nourl}

\appendix
\section{Data Statistics}\label{app:data}

\setlength{\tabcolsep}{1pt}

\begin{table}[h]
\centering
{\small
\begin{tabular}{c|c c c | c}
& \multicolumn{3}{c}{\textbf{Europarl}} & \multicolumn{1}{|c}{\textbf{Subtitles}}\\
\cline{2-5}
& \textbf{En-Fr} & \textbf{En-Et} & \textbf{En-De} & \textbf{En-Ru} \\
\hline
\hline
\multicolumn{5}{c}{\textbf{Dev/Test}}\\
\hline
{\# Conversations} & {140/209} & {88/132} & 70/108 & 462/694 \\
{\# Sentences} & {4.9k/7.8k} & {3.2k/5.2k} & 2.1k/3.3k & 5.9k/9k \\
\hline
\end{tabular}
}
\caption{General statistics for development and test sets. }
\label{tab:auxdata}
\end{table}

\section{Experiments}\label{app:exp}

\paragraph{Training}

For the base model, we make use of stochastic gradient descent (SGD)\footnote{In our preliminary experiments, we tried SGD, Adam and Adagrad as optimisers, and found SGD to achieve better perplexities in lesser number of epochs \cite{Bahar:17}.} with initial learning rate of 0.1 and a decay factor of 0.5 after the fifth epoch for a total of 15 epochs. For the contextual model, we use SGD with an initial learning rate of 0.08 and  a decay factor of 0.9 after the first epoch for a total of 30 epochs. To avoid overfitting, we employ dropout and set its rate to 0.2. To reduce the training time of our contextual model, we perform computation of one turn at a time, for instance, when using the source context, we run the Turn-RNNs for previous turns once and re-run the Turn-RNN only for sentences in the current turn.

\end{document}